\documentclass[runningheads]{llncs}
\usepackage{graphicx}
\usepackage{amsmath,amssymb} % define this before the line numbering.
\usepackage{color}
\usepackage[width=122mm,left=12mm,paperwidth=146mm,height=193mm,top=12mm,paperheight=217mm]{geometry}

\usepackage{wasysym}
\usepackage[ruled,vlined]{algorithm2e}
\usepackage{multirow}

\begin{document}
% \linenumbers

% Commands added by me.
\newcommand{\bigcell}[2]{\begin{tabular}{@{}#1@{}}#2\end{tabular}}
\newcommand{\Aref}[1]{Algorithm~\ref{#1}}
\newcommand{\Tref}[1]{Table~\ref{#1}}
\newcommand{\Eref}[1]{Eq.~(\ref{#1})}
\newcommand{\Fref}[1]{Fig.~\ref{#1}}
\newcommand{\Sref}[1]{Sec.~\ref{#1}}
\newcommand{\etal}[0]{\textit{et al.}}

\pagestyle{headings}
\mainmatter

\title{Pixel-Level Domain Transfer} % Replace with your title

\titlerunning{Pixel-Level Domain Transfer}

\authorrunning{Yoo \textit{et al}.}

\author{\small Donggeun Yoo$^1$, Namil Kim$^1$, Sunggyun Park$^1$, Anthony S. Paek$^2$, In So Kweon$^1$}

%Please write out author names in full in the paper, i.e. full given and family names. 
%If any authors have names that can be parsed into FirstName LastName in multiple ways, please include the correct parsing, in a comment to the volume editors:
%\index{Lastnames, Firstnames}
%(Do not uncomment it, because you may introduce extra index items if you do that...)

%\institute{$^1$KAIST, Daejeon, South Korea. $^2$Lunit Inc., Seoul, South Korea.\\
%\email{\{dgyoo,nikim\}@rcv.kaist.ac.kr \{sunggyun,iskweon\}@kaist.ac.kr apaek@lunit.io}
% \email{ \{dgyoo@rcv.,nikim@rcv.,sunggyun@,iskweon@\}kaist.ac.kr apaek@lunit.io}
%}
\institute{$^1$KAIST, Daejeon, South Korea.\\
$^2$Lunit Inc., Seoul, South Korea.\\
\email{\{dgyoo,nikim\}@rcv.kaist.ac.kr\\\{sunggyun,iskweon\}@kaist.ac.kr}\\
\email{apaek@lunit.io}\\
}

\maketitle

\begin{abstract}
We present an image-conditional image generation model. The model transfers an input domain to a target domain in semantic level, and generates the target image in pixel level. To generate realistic target images, we employ the real/fake-discriminator as in Generative Adversarial Nets \cite{goodfellow14}, but also introduce a novel domain-discriminator to make the generated image relevant to the input image. We verify our model through a challenging task of generating a piece of clothing from an input image of a dressed person. We present a high quality clothing dataset containing the two domains, and succeed in demonstrating decent results.

\end{abstract}

%% INTRODUCTION
\section{Introduction}
Every morning, we agonize in front of the closet over what to wear, how to dress up, and imagine ourselves with different clothes on. To generate mental images~\cite{eysenck06} of ourselves wearing clothes on a hanger is an effortless work for our brain. In our daily lives, we ceaselessly perceive visual scene or objects, and often transfer them to different forms by the mental imagery. 
Our focus of this paper lies on the problem; to enable a machine to transfer a visual input into different forms and to visualize the various forms by generating a pixel-level image.

Image generation has been attempted by a long line of works~\cite{hinton06,salakhutdinov09,vincent08} but generating realistic images has been challenging since an image itself is high dimensional and has complex relations between pixels. However, several recent works have succeeded in generating realistic images~\cite{goodfellow14,gregor15,sohl15,theis15}, with the drastic advances of deep learning. Although these works are similar to ours in terms of image generation, ours is distinct in terms of \textit{image-conditioned image generation}. We take an image as a conditioned input lying in a domain, and re-draw a target image lying on another.

In this work, we define two domains; a source domain and a target domain. The two domains are connected by a semantic meaning. For instance, if we define an image of a dressed person as a source domain, a piece of the person's clothing is defined as the target domain. Transferring an image domain into a different image domain has been proposed in computer vision~\cite{saenko10,kulis11,gopalan11,oquab14,chen15,huang15}, but all these adaptations take place in the feature space, i.e. the model parameters are adapted. However, our method directly produces target images.

\begin{figure}[t]
\centering
\begin{tabular}{ccccc}
\includegraphics[width=0.17\linewidth]{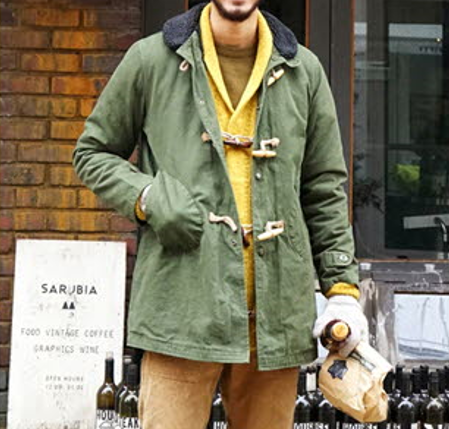}&
\includegraphics[width=0.18\linewidth]{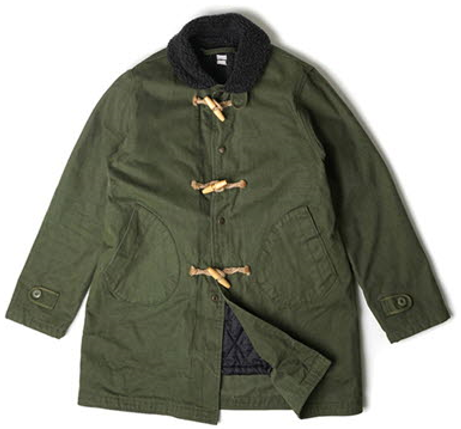}&
\includegraphics[width=0.19\linewidth]{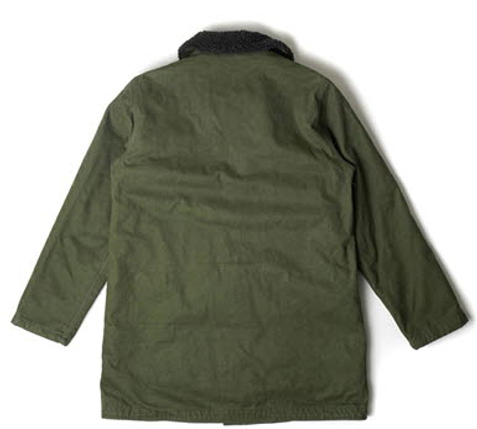}&
\includegraphics[width=0.20\linewidth]{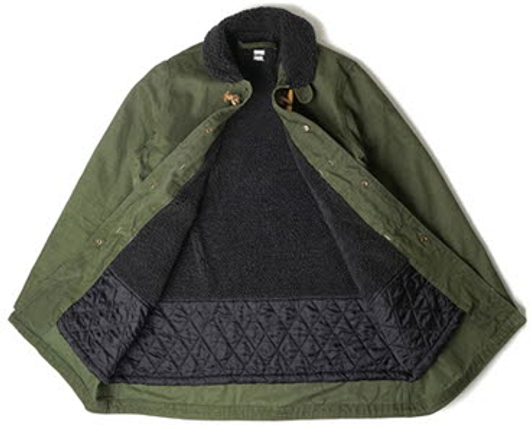}&
\includegraphics[width=0.18\linewidth]{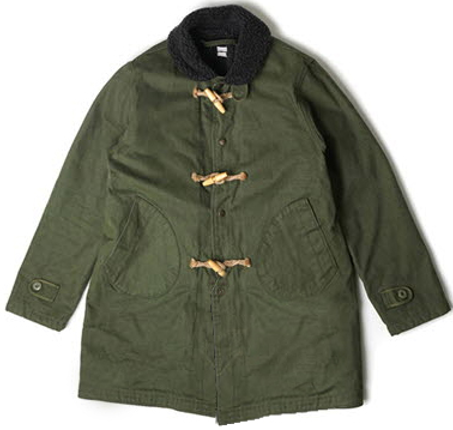}\\
A source image.&\multicolumn{4}{c}{Possible target images.}
\end{tabular}
\caption{A real example showing non-deterministic property of target image in the pixel-level domain transfer problem.}
\label{fig:nonunique}
\end{figure}

We transfer a knowledge in a source domain to a pixel-level target image while overcoming the semantic gap between the two domains. Transferred image should look realistic yet preserving the semantic meaning. To do so, we present a pixel-level domain converter composed of an encoder for semantic embedding of a source and a decoder to produce a target image. However, training the converter is not straightforward because the target is not deterministic~\cite{yan15}. Given a source image, the number of possible targets is unlimited as the examples in \Fref{fig:nonunique} show. To challenge this problem, we introduce two strategies as follows.

To train our converter, we first place a separate network named \textit{domain discriminator} on top of the converter. The domain discriminator takes a pair of a source image and a target image, and is trained to make a binary decision whether the input pair is associated or not. The domain discriminator then supervises the converter to produce associated images. Both of the networks are jointly optimized by the adversarial training method, which Goodfellow~\etal~\cite{goodfellow14} propose for generating realistic images. Such binary supervision solves the problem of non-deterministic property of the target domain and enables us to train the semantic relation between the domains. Secondly, in addition to the domain discriminator, we also employ the discriminator of~\cite{goodfellow14}, which is supervised by the labels of ``real'' or ``fake'', to produce realistic images.

Our framework deals with the three networks that play distinct roles. Labels are given to the two discriminators, and they supervise the converter to produce images that are realistic yet keeping the semantic meaning. Those two discriminators become unnecessary after the training stage and the converter is our ultimate goal. We verify our method by quite challenging settings; the source domain is a natural human image and the target domain is a product image of the person's top. To do so, we have made a large dataset named LookBook, which contains in total of 84k images, where 75k human images are associated with 10k top product images. With this dataset, our model succeeds in generating decent target images, and the evaluation result verifies the effectiveness of our \textit{domain discriminator} to train the converter.
\clearpage
\subsubsection{Contributions}
In summary, our contributions are,
\begin{enumerate}
\item Proposing the first framework for semantically transferring a source domain to a target domain in pixel-level.
\item Proposing a novel discriminator that enables us to train the semantic relation between the domains.
\item Building a large clothing dataset containing two domains, which is expected to contribute to a wide range of domain adaptation researches.
\end{enumerate}

\section{Related Work}
Our work is highly related with the image-generative models since our final result from an input image is also an image. The image-generative models can be grouped into two families; generative parametric approaches~\cite{hinton06,salakhutdinov09,vincent08} and adversarial approaches~\cite{goodfellow14,mirza14,Denton15,Radford15}. The generative parametric approaches often have troubles in training complexities, which results in a low rate of success in generating realistic natural images. The adversarial approaches originate from Generative Adversarial Nets (GAN) proposed by Goodfellow~\etal~\cite{goodfellow14}. GAN framework introduces a generator (i.e. a decoder), which generates images, and a discriminator, which distinguishes between generated samples and real images. The two networks are optimized to go against each other; the discriminator is trained to distinguish between real and fake samples while the generator is trained to confuse the discriminator. Mirza and Osindero~\cite{mirza14} extend GAN to a class conditional version, and Denton~\etal~\cite{Denton15} improve the image resolution in a coarse-to-fine fashion. However, GAN is known to be unstable due to the adversarial training, often resulting in incomprehensible or noisy images. Quite recently, Radford~\etal~\cite{Radford15} have proposed architectures named Deep Convolutional GANs, which is relatively more stable to be trained, and have succeeded in generating high quality images. As approaches focusing on different network architectures, a recurrent network based model~\cite{gregor15} and a deconvolutional network based model~\cite{dosovitskiy15} have also been proposed.

The recent improvements of GAN framework and its successful results motivate us to adopt the networks. We replace the generator with our converter which is an image-conditioned model, while \cite{mirza14} is class-conditional and \cite{yan15} is attribute-conditional. The generator of Mathieu~\etal~\cite{mathieu15} is similar to ours in that it is conditioned with video frames to produce next frames. They add a mean square loss to the generator to strongly relate the input frames to the next frames. However, we cannot use such loss due to the non-deterministic property of the target domain. We therefore introduce a novel discriminator named domain discriminator. 

Our work is also related with the transfer learning, also called as the domain adaptation. This aims to transfer the model parameter trained on a source domain to a different domain. For visual recognition, many methods to adapt domains \cite{saenko10,kulis11,gopalan11} have been proposed. Especially for the recent use of the deep convolutional neural network~\cite{lecun89}, it has been common to pre-train a large network~\cite{krizhevsky12} over ImageNet~\cite{russakovsky15} and transfer the parameters to a target domain~\cite{oquab14,razavian14,yoo15}. Similar to our clothing domains, Chen~\etal~\cite{chen15} and Huang~\etal~\cite{huang15} address a gap between fashion shopping mall images and unconstrained human images for the clothing attribute recognition~\cite{chen15} and the product retrieval~\cite{huang15}. Ganin and Lempitsky~\cite{ganin15} also learns domain-invariant features by the adversarial training method. However, all these methods are different from ours in respect of cross-domain \textit{image generation}. The adaptation of these works takes place in the feature space, while we directly produce target images from the source images.

\section{Review of Generative Adversarial Nets}
\label{sec:gan}
Generative Adversarial Nets (GAN) \cite{goodfellow14} is a generalized framework for generative models which \cite{Denton15,Radford15,mathieu15} and we utilize for visual data. In this section, we briefly review GAN in the context of image data. GAN is formed by an adversarial setting of two networks; a generator and a discriminator. The eventual goal of the generator is to map a small dimensional space $Z$ to a pixel-level image space, i.e., to enable the generator to produce a realistic image from an input random vector $z\in Z$.

To train such a generator, a discriminator is introduced. The discriminator takes either a real image or a fake image drawn by the generator, and distinguishes whether its input is real or fake. The training procedure can be intuitively described as follows. Given an initialized generator $G^0$, an initial discriminator $D_R^0$ is firstly trained with real training images $\{I^i\}$ and fake images $\{\hat{I}^j=G^0(z^j)\}$ drawn by the generator. After that, we freeze the updated discriminator $D_R^1$ and train the generator $G^0$ to produce better images, which would lead the discriminator $D_R^1$ to misjudge as real images. These two procedures are repeated until they converge. The objective function can be represented as a minimax objective as,
\begin{equation}
\min_{\Theta^G}\max_{\Theta_R^D}\mathbb{E}_{I\sim p_{\text{data}}(\mathbf{I})}[\log(D_R(I))]+
\mathbb{E}_{z\sim p_{\text{noise}}(\mathbf{z})}[\log(1-D_R(\hat{I}))],
\end{equation}
where $\Theta^G$ and $\Theta_R^D$ indicate the model parameters of the generator and the discriminator respectively. Here, the discriminator produces a scalar probability that is high when the input $I$ is real but otherwise low. The discriminator loss function $\mathcal{L}_R^D$ is defined as the binary cross entropy,
\label{eq:gan}
\begin{multline}
\mathcal{L}_R^D\left(I\right)=-t\cdot \log[D_R(I)] + (t-1)\cdot\log[1-D_R(I)],\\
\text{s.t.}\;\;t=\left\{
\begin{matrix}
1&\;\;\text{if}\;\;I\in\{I^i\}\\
0&\;\;\;\text{if}\;\;I\in\{\hat{I}^j\}.
\end{matrix}\right.
\label{eq:d}
\end{multline}

One interesting fact in the GAN framework is that the model is trained under the lowest level of supervision; real or fake. Without strong and fine supervisions (e.g. mean square error between images), this framework succeeds in generating realistic images. This motivates us to raise the following question. 
Under such a low-level supervision, would it be possible to train a connection between distinct image domains? If so, could we transform an image lying in a domain to a realistic image lying on another? Through this study, we have succeeded in doing so, and the method is to be presented in \Sref{sec:method}.

%% METHOD
\section{Pixel-Level Domain Transfer}
\label{sec:method}
In this section, we introduce the pixel-level domain transfer problem. Let us define a source image domain $S\subset \mathbb{R}^{W\times H\times3}$ and a target image domain $T\subset \mathbb{R}^{W\times H\times3}$. Given a transfer function named a converter $C$, our task is to transfer a source image $I_S\in S$ to a target image $\hat{I}_T\in T$ such as
\begin{equation}
\hat{I}_T=C(I_S|\Theta^C),
\end{equation}
where $\Theta^C$ is the model parameter of the converter. Note that the inference $\hat{I}_T$ is not a feature vector but itself a target image of $W\times H\times3$ size. To do so, we employ a convolutional network model for the converter $C$, and adopt a supervised learning to optimize the model parameter $\Theta^C$. In the training data, each source image $I_S$ should be associated with a ground-truth target image $I_T$.

\subsection{Converter Network}
\label{ssec:converter}
Our target output is a \textit{pixel-level} image. Furthermore, the two domains are connected by a \textit{semantic} meaning. Pixel-level generation itself is challenging but the semantic transfer makes the problem even more difficult. A converter should selectively summarize the semantic attributes from a source image and then produce a transformed pixel-level image.

\begin{figure}[t!]
\centering
\includegraphics[width=1\linewidth]{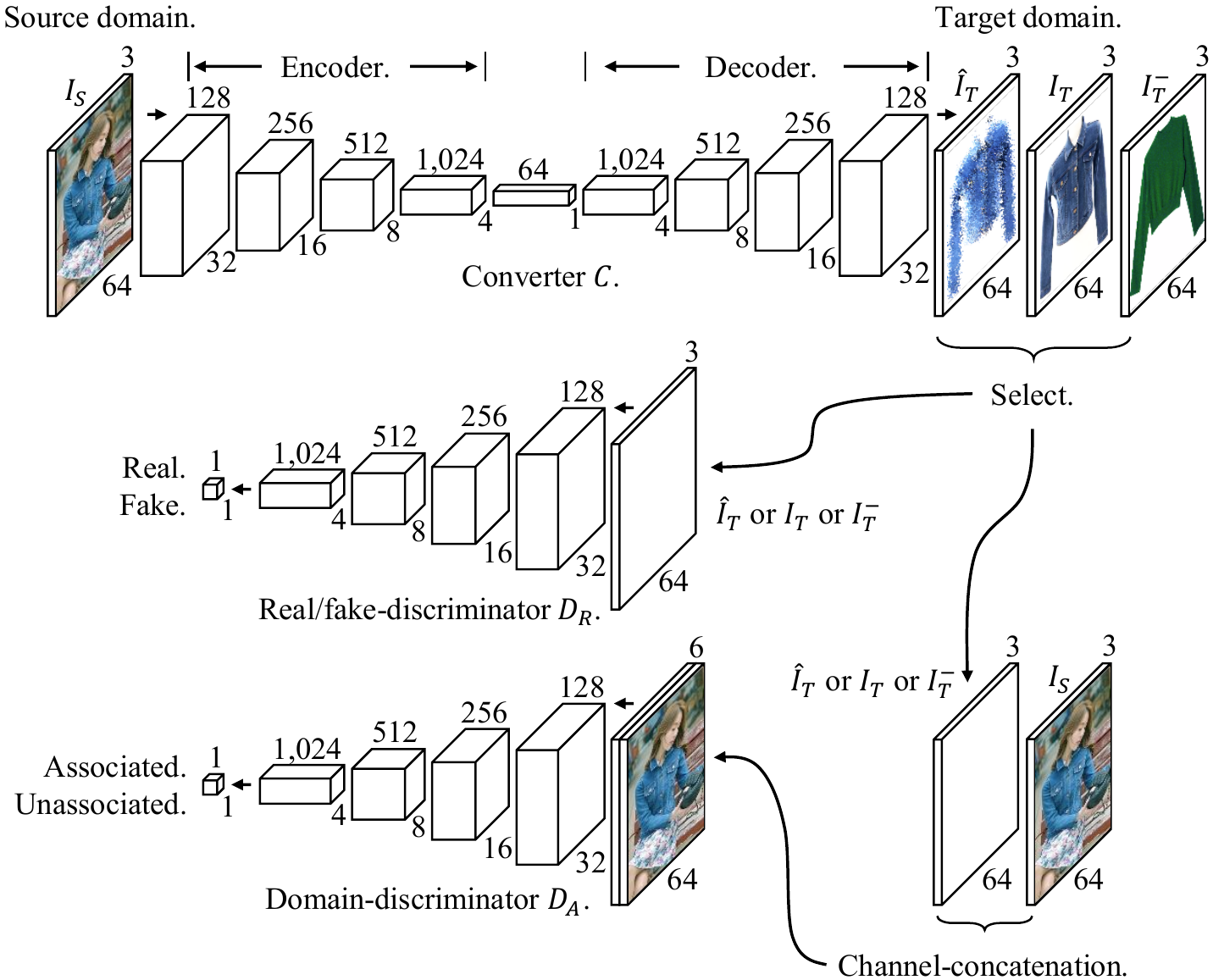}
\caption{Whole architecture for pixel-level domain transfer.}
\label{fig:arch}
\end{figure}
\begin{table}[t]
\centering
\begin{tabular}{|c|c|c|c|c|c|c|}\hline
Layer&\bigcell{c}{Number\\of filters}&\bigcell{c}{Filter size\\(w$\times$h$\times$ch)}&Stride&Pad&\bigcell{c}{Batch\\norm.}&\bigcell{c}{Activation\\function}\\\hline\hline
Conv. 1&128&5$\times$5$\times$\{3, 3, 6\}&2&2&$\times$&L-ReLU\\
Conv. 2&256&5$\times$5$\times$128&2&2&$\ocircle$&L-ReLU\\
Conv. 3&512&5$\times$5$\times$256&2&2&$\ocircle$&L-ReLU\\
Conv. 4&1,024&5$\times$5$\times$512&2&2&$\ocircle$&L-ReLU\\
Conv. 5&\{64, 1, 1\}&1$\times$1$\times$1,024&1&0&$\{\ocircle,\times,\times\}$&\{L-ReLU, sigmoid, sigmoid\}\\\hline
\end{tabular}\\
(a) Details of the \{encoder, real/fake discriminator, domain discriminator\}.\\$\;$\\
\begin{tabular}{|c|c|c|c|c|c|c|}\hline
Layer&\bigcell{c}{Number\\of filters}&\bigcell{c}{Filter size\\(w$\times$h$\times$ch)}&Stride&Pad&\bigcell{c}{Batch\\norm.}&\bigcell{c}{Activation\\function}\\\hline\hline
Conv. 1&4$\times$4$\times$1,024&1$\times$1$\times$64&1&0&$\ocircle$&ReLU\\
F-Conv. 2&1,024&5$\times$5$\times$512&1/2&-&$\ocircle$&ReLU\\
F-Conv. 3&512&5$\times$5$\times$256&1/2&-&$\ocircle$&ReLU\\
F-Conv. 4&256&5$\times$5$\times$128&1/2&-&$\ocircle$&ReLU\\
F-Conv. 5&128&5$\times$5$\times$3&1/2&-&$\times$&tanh\\\hline
\end{tabular}\\
(b) Details of the decoder.\\$\;$\\
\caption{Details of each network. In (a), each entry in \{$\cdot$\} corresponds to each network. L-ReLU is leaky-ReLU. In (b), F denotes fractional-stride. The activation from the first layer is reshaped into 4$\times$4$\times$1,024 size before being fed to the second layer.}
\label{tab:arch}
\end{table}

The top network in \Fref{fig:arch} shows the architecture of the converter we propose. The converter is a unified network that is end-to-end trainable but we can divide it into the two parts; an encoder and a decoder. The encoder part is composed of five convolutional layers to abstract the source into a semantic 64-dimensional code. This abstraction procedure is significant since our source domain (e.g. natural fashion image) and target domain (e.g. product image) are paired in a semantic content (e.g. the product). The 64-dimensional code should capture the semantic attributes (e.g. category, color, etc.) of a source to be well decoded into a target. The code is then fed by the decoder, which constructs a relevant target through the five decoding layers. Each decoding layer conducts the fractional-strided convolutions, where the convolution operates in the opposite direction. The reader is referred to \Tref{tab:arch} for more details about the architectures of the encoder and the decoder.

\subsection{Discriminator Networks}
\label{ssec:discrim}

Given the converter, a simple choice of a loss function to train it is the mean-square error (MSE) such as $||\hat{I}_T-I_T||_2^2$. However, MSE may not be a proper choice due to critical mismatches between MSE and our problem. Firstly, MSE is not suitable for pixel-level supervision for natural images. It has been well known that MSE is prone to produce blurry images because it inherently assumes that the pixels are drawn from Gaussian distribution \cite{mathieu15}. Pixels in natural images are actually drawn from complex multi-modal distributions. Besides its intrinsic limitation, it causes another critical problem especially for the pixel-level domain transfer as follows.

Given a source image, the target is actually not unique in our problem. Our target domain is the lowest pixel-level image space, not the high-level semantic feature space. Thus, the number of possible targets from a source is infinite. \Fref{fig:nonunique} is a typical example showing that the target is not unique. The clothing in the target domain is captured in various shapes, and all of the targets are true. Besides the shapes, the target image can be captured from various viewpoints, which results in geometric transformations. However, minimizing MSE always forces the converter to fit into one of them. Image-to-image training with MSE never allows a small geometric miss-alignment as well as various shapes. Thus, training the converter with MSE is not a proper use for this problem. It would be better to introduce a new loss function which is tolerant to the diversity of the pixel-level target domain.

In this paper, on top of the converter, we place a discriminator network which plays a role as a loss function. As in \cite{goodfellow14,Denton15,Radford15}, the discriminator network guides the converter to produce realistic target under the supervision of real/fake. However, this is not the only role that our discriminator plays. If we simply use the original discriminator replacing MSE, a produced target could look realistic but its contents may not be relevant to the source. This is because there is no pairwise supervision such as MSE. Only real/fake supervision exists. 

Given arbitrary image triplets ($I_S^+, I_S^{\oplus}, I_S^-$) in the source domain $S$, where $I_S^+$ and $I_S^{\oplus}$ are about the same object while $I_S^-$ is not, a converter transfers them into the images ($\hat{I}_T^+, \hat{I}_T^{\oplus}, \hat{I}_T^-$) in the target domain $T$. Let us assume that these transferred images look realistic due to the real/fake discriminator. Beyond the realistic results, the best converter $C$ should satisfy the following condition,
\begin{equation}
s\left(\hat{I}_T^+, \hat{I}_T^{\oplus}\right) > s\left(\hat{I}_T^+, \hat{I}_T^-\right) \;\;\text{and}\;\; s\left(\hat{I}_T^+, \hat{I}_T^{\oplus}\right) > s\left(\hat{I}_T^{\oplus}, \hat{I}_T^-\right),
\label{eq:inequality}
\end{equation}
where $s$($\cdot$) is a semantic similarity function. This condition means that an estimated target should be semantically associated with the source. One supervision candidate to let the converter $C$ meet the condition is the combined use of MSE with the real/fake loss. However, again, it is not the best option for our problem because the ground-truth $I_T$ is not unique. Thus, we propose a novel discriminator, named domain discriminator, to take the pairwise supervision into consideration.

The domain discriminator $D_A$ is the lowest network illustrated in \Fref{fig:arch}. To enable pairwise supervision while being tolerant to the target diversity, we significantly loosen the level of supervision compared to MSE. The network $D_A$ takes a pair of source and target as input, and produces a scalar probability of whether the input pair is associated or not. Let us assume that we have a source $I_S$, its ground truth target $I_T$ and an irrelevant target $I_T^-$. We also have an inference $\hat{I}_T$ from the converter $C$.
We then define the loss $\mathcal{L}_A^D$ of the domain discriminator $D_A$ as,
\begin{multline}
\mathcal{L}_A^D(I_S,I)=-t\cdot \log[D_A(I_S, I)] + (t-1)\cdot\log[1-D_A(I_S, I)],\\
\text{s.t.}\;\;t=\left\{
\begin{matrix}
1&\;\text{if}\;\;I=I_T\\
0&\;\text{if}\;\;I=\hat{I}_T\\
0&\;\;\text{if}\;\;I=I_T^-.
\end{matrix}\right.
\label{eq:dd}
\end{multline}
The source $I_S$ is always fed by the network as one of the input pair while the other $I$ is chosen among ($I_T^-$, $\hat{I}_T$, $I_T$) with equal probability. Only when the source $I_S$ and its ground-truth $I_T$ is paired as input, the domain discriminator is trained to produce high probability whereas it minimizes the probability in other cases. Here, let us pay more attention to the input case of ($I_S$, $\hat{I}_T$).

The produced target $\hat{I}_T$ comes from the source but we regard it as an unassociated pair ($t$=0) when we train the domain discriminator. Our intention of doing so is for \textit{adversarial training} of the converter and the domain discriminator. The domain discriminator loss is minimized for training the domain discriminator while it is maximized for training the converter. The better the domain discriminator distinguishes a ground-truth $I_T$ and an inference $\hat{I}_T$, the better the converter transfers the source into a relevant target.

In summary, we employ both of the real/fake discriminator and the domain discriminator for adversarial training. These two networks play a role as a loss to optimize the converter, but have different objectives. The real/fake discriminator penalizes an unrealistic target while the domain discriminator penalizes a target being irrelevant to a source. The architecture of the real/fake discriminator is identical to that of \cite{Radford15} as illustrated in \Fref{fig:arch}. The domain discriminator also has the same architecture except for the input filter size since our input pair is stacked across the channel axis. Several architecture families have been proposed to feed a pair of images to compare them but a simple stack across the channel axis has shown the best performance as studied in \cite{zagoruyko15}. The reader is referred to \Tref{tab:arch} for more details about the discriminator architectures.

\subsection{Adversarial Training}
\label{ssec:training}
In this section, we present the method for training the converter $C$, the real/fake discriminator $D_R$ and the domain discriminator $D_A$. Because we have the two discriminators, the two loss functions have been defined. The real/fake discriminator loss $\mathcal{L}_R^D$ is \Eref{eq:d}, and the domain discriminator loss $\mathcal{L}_A^D$ is \Eref{eq:dd}. With the two loss functions, we follow the adversarial training procedure of \cite{goodfellow14}. 

Given a paired image set for training, let us assume that we get a source batch $\{I_S^i\}$ and a target batch $\{I^i\}$ where a target sample $I^i$ is stochastically chosen from $(I_T^i, I_T^{i-}, \hat{I}_T^i)$ with an equal probability. At first, we train the discriminators.
We train the real/fake discriminator $D_R$ with the target batch to reduce the loss of \Eref{eq:d}. The domain discriminator $D_A$ is trained with both of source and target batches to reduce the loss of \Eref{eq:dd}. 
After that, we freeze the updated discriminator parameters $\{\hat{\Theta}_R^D,\hat{\Theta}_A^D\}$, and optimize the converter parameters $\Theta^C$ to \textit{increase} the losses of both discriminators. 
The loss function of the converter can be represented as,
\begin{equation}
\mathcal{L}^C(I_S,I)=-\frac{1}{2}\mathcal{L}_R^D\left(I\right)-\frac{1}{2}\mathcal{L}_A^D(I_S,I),\;\;\text{s.t.}\;\;I=\text{sel}\left(\{I_T,\hat{I}_T,I_T^-\}\right),
\end{equation}
where sel($\cdot$) is a random selection function with equal probability. The reader is referred to \Aref{alg:train} for more details of the training procedures.
%Given a paired image set, let us assume we have a triplets $\left(I_S, I_T, I_T^-\right)$ and an inference $\hat{I}_T$, where $I_T^-$ is a target irrelevant to $I_S$ randomly selected from the set. At first, we train the real/fake discriminator $D_R$
%

\begin{algorithm}[t]
\SetAlgoLined
\textbf{Set} the learning rate $\eta$ and the batch size $B$.\\
\textbf{Initialize} each network parameters $\Theta^C, \Theta_R^D, \Theta_A^D$,\\
\KwData{Paired image set $\{I_S^n, I_T^n\}_{n=1}^N$.}
\While{not converged}{
Get a source batch $\left\{I_S^i\right\}_{i=1}^B$ and a target batch $\left\{I^i\right\}_{i=1}^B$,\\
$\;\;\;\;\;\;$where $I^{i}$ is a target sample randomly chosen from $(I_T^i, I_T^{i-}, \hat{I}_T^i)$.\\
\textbf{Update the real/fake discriminator $D_R$:}\\
$\;\;\;\;\;\;\Theta_R^D\leftarrow\Theta_R^D-\eta\cdot\frac{1}{B}\sum_{i=1}^B\frac{\partial\mathcal{L}_R^D\left(I^i\right)}{\partial\Theta_R^D}$\\
\textbf{Update the domain discriminator $D_A$:}\\
$\;\;\;\;\;\;\Theta_A^D\leftarrow\Theta_A^D-\eta\cdot\frac{1}{B}\sum_{i=1}^B\frac{\partial\mathcal{L}_A^D\left(I_S^i,I^i\right)}{\partial\Theta_A^D}$\\
\textbf{Update the converter $C$:}\\
$\;\;\;\;\;\;\Theta^C\leftarrow\Theta^C-\eta\cdot\frac{1}{B}\sum_{i=1}^B\frac{\partial\mathcal{L}^C\left(I_S^i,I^i\right)}{\partial\Theta^C}$\\
}
\caption{Adversarial training for the pixel-level domain transfer.}
\label{alg:train}
\end{algorithm}

\section{Evaluation}
In this section, we verify our pixel-level domain transfer by a challenging task; a natural human image belongs to the source domain, and a product image of that person's top belongs to the target domain. We first give a description on the dataset in~\Sref{sec:db}. We then provide details on the experimental setting in \Sref{sec:setting}, and we demonstrate and discuss the results in \Sref{sec:quali}$\sim$\ref{sec:quanti2}.

\subsection{LookBook Dataset}
\label{sec:db}
\begin{figure}[t]
\centering
\includegraphics[width=1\linewidth]{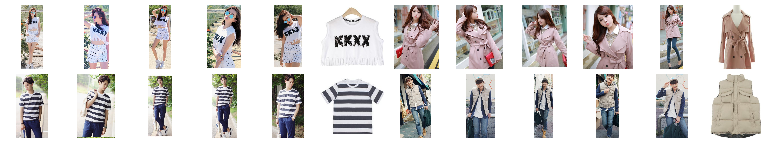}
\caption{Example images of LookBook. A product image is associated with multiple fashion model images.}
\label{fig:lookbook}
\end{figure}

We make a dataset named LookBook that covers two fashion domains. Images of one domain contain fashion models, and those of the other domain contain top products with a clean background. Real examples are shown in \Fref{fig:lookbook}. We manually associate each product image with corresponding images of a fashion model fitting the product, so each pair is accurately connected with the same product. LookBook contains 84,748 images where 9,732 top product images are associated with 75,016 fashion model images. It means that a product has around 8 fashion model images in average. We collect the images from five on-line fashion shopping malls\footnote{\{bongjashop, jogunshop, stylenanda\}.com, \{smallman, wonderplace\}.co.kr} where a product image and its fashion model images are provided. Although we utilize LookBook for the pixel-level domain transfer, we believe that it can contribute to a wide range of domain adaptation researches.

Chen~\etal~\cite{chen15} also has presented a similar fashion dataset dealing with two domains. However, it is not suitable for our task since the domains are differently defined in details. They separate the domain into user taken images and on-line shopping mall images so that both domains include humans.

\subsection{Experiment Details}
\label{sec:setting}
Before training, we rescale all images in LookBook to have 64 pixels at a longer side while keeping the aspect ratio, and fill the margins of both ends with 255s. Pixels are normalized to a range of $[-1, 1]$ according to the tanh activation layer of the converter. We then randomly select 5\% images to define a validation set, and also 5\% images for a test set. Since LookBook has 9,732 products, each of the validation set and the test set is composed of 487 product images and their fashion model images. The remaining images compose a training set.

The filters of the three networks are randomly initialized from a zero mean Gaussian distribution with a standard deviation of 0.02. The leak slope of the LeakyReLU in~\Tref{tab:arch}-(a) is 0.2. All models were trained with Stochastic Gradient Descent with mini-batch of 128 size. We also follow the learning rate of 0.0002 and the momentum of 0.5 suggested by \cite{Radford15}. After 25 epochs, we lessen the learning rate to 0.00002 for 5 more epochs.

\Tref{tab:notation} shows the notations and the descriptions of the 4 baselines and our method. The training details of all the baselines are identical to those of ours.

\subsection{Qualitative evaluation}
\label{sec:quali}
First, we show qualitative results in~\Fref{fig:comp}, where the examples are chosen from the test set. Our results look more relevant to the source image and more realistic compared to those of baselines. Boundaries of products are sharp, and small details such as stripes, patterns are well described in general. The results of ``C+RF'' look realistic but irrelevant to the source image, and those of ``C+MSE'' are quite blurry.

\begin{figure}[t]
\centering
\includegraphics[width=1\linewidth]{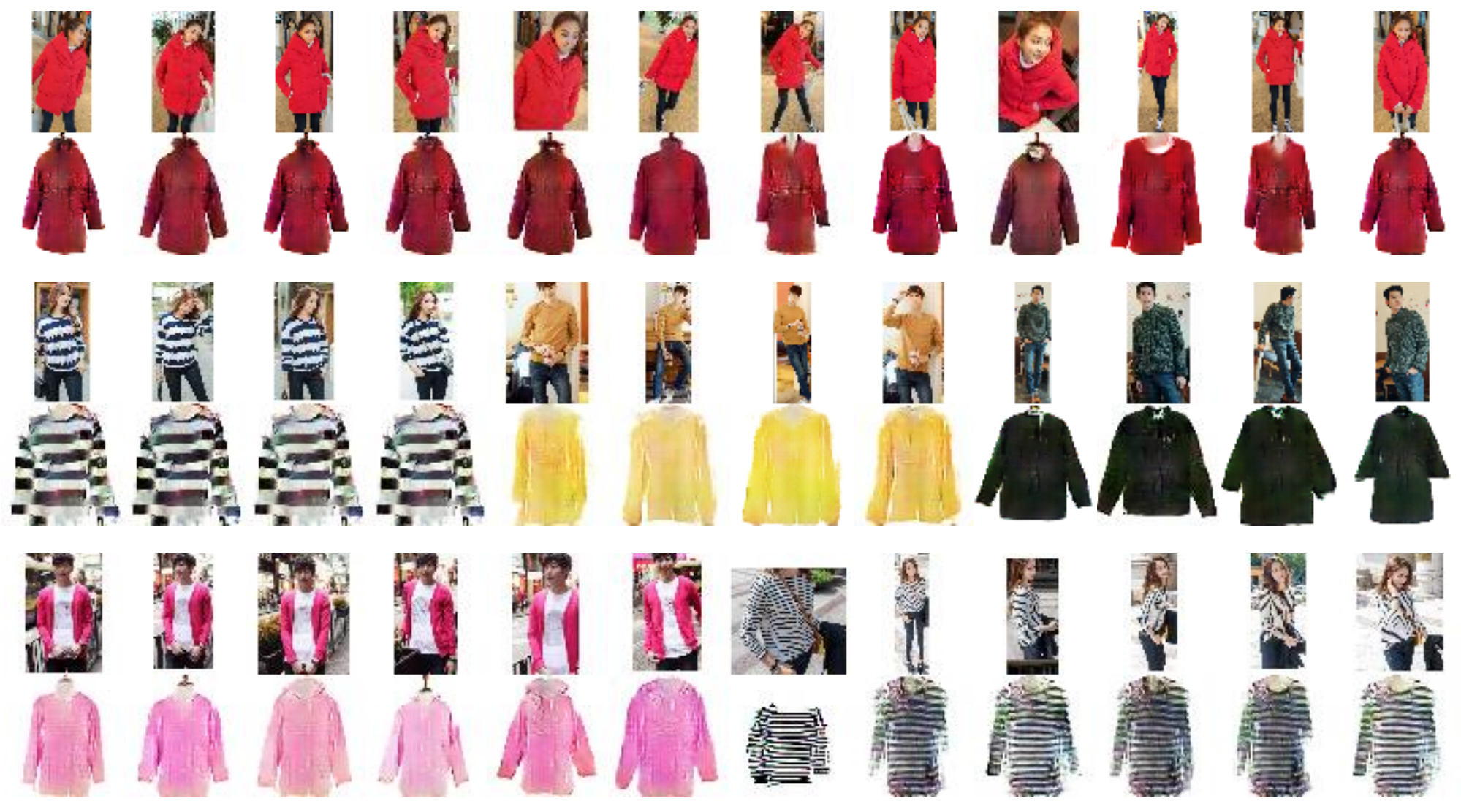}
\caption{Generation results under varying input conditions. The odd rows are inputs, and the even rows are generation results. Each image is in 64$\times$64$\times$3 dimensions.}
\label{fig:invariance}
\end{figure}

\Fref{fig:invariance} verifies how well the encoder of the converter encodes clothing attributes under the various conditions of source images. The source images significantly vary in terms of backgrounds, viewpoints, human poses and self-occlusions. Despite these variations, our converter generates less varying targets while reflecting the clothing attributes and categories of the source images. These results imply that the encoder robustly summarizes the source information in a semantic level.

\begin{table}[t]
\begin{center}
\begin{tabular}{|l|l|}\hline
Notations&Descriptions\\\hline\hline
C+RF&A converter trained only with the real/fake discriminator.\\\hline
C+MSE&A converter trained only with the mean square loss.\\\hline
C+RF+DD$-$Neg&\bigcell{l}{A converter trained with both of the discriminators.\\Negative pairs are not used. Only positive pairs are used.}\\\hline
Retrieval by DD-score&\bigcell{l}{Retrieving the nearest product image in the training set.\\The queries are the human images in the test set.\\The retrieval scores come from the domain discriminator.}\\\hline
C+RF+DD (Ours)&A converter trained with both of the discriminators.\\\hline
\end{tabular}
\end{center}
\caption{Notations and descriptions of baselines and our method.}
\label{tab:notation}
\end{table}
\begin{table}[t]
\begin{center}
\begin{tabular}{|l|c|c|c||l|c|c|}\hline
\multicolumn{4}{|c||}{User study score}&\multicolumn{3}{c|}{Pixel-level (dis)similarity}\\\hline
Methods&Real&Att&Cat&Methods&RMSE&C-SSIM\\\hline\hline
C+RF&0.40&0.21&0.06&C+RF&0.39&0.18\\
C+MSE&0.28&0.60&0.60&C+MSE&\textbf{0.26}&0.20\\
C+RF+DD (Ours)&\textbf{0.82}&\textbf{0.67}&\textbf{0.77}&C+RF+DD$-$Neg&0.32&0.18\\
&&&&Retrieval by DD-score&0.44&0.19\\
&&&&C+RF+DD (Ours)&0.32&\textbf{0.21}\\\hline
\end{tabular}
\end{center}
\caption{Quantitative evaluations. All the values are normalized to a range of $[0, 1]$.}
\label{tab:eval}
\end{table}

\subsection{Quantitative evaluation by user study}
\label{sec:quanti1}
Since the target domain is not deterministic, it is difficult to quantitatively analyze the performance. Thus, we conduct a user study on our generation results as a primary evaluation. We compare our method with the top two baselines in \Tref{tab:notation}. For this study, we created a sub-test set composed of 100 source images randomly chosen from the test set. For each source image, we showed users three target images generated by the two baselines and our method. Users were asked to rate them three times in accordance with three different evaluation criteria as follows. A total of 25 users participated in this study.
\begin{enumerate}
\item How realistic is each result? Give a score from 0 to 2.
\item How well does each result capture the attributes (color, texture, logos, etc.) of the source image? Give a score from 0 to 2.
\item Is the category of each result identical to that of the source image? Give a binary score of 0 or 1.
\end{enumerate}

\begin{figure}[h!]
\centering
{\small Source$\;\;\;$RF$\;\;\;\;$MSE$\;\;\;$Ours$\;\;\;\;\;\;$Source$\;\;\;$RF$\;\;\;\;$MSE$\;\;\;$Ours$\;\;\;\;\;\;$Source$\;\;\;$RF$\;\;\;\;$MSE$\;\;\;$Ours$\;$}
\includegraphics[width=1\linewidth]{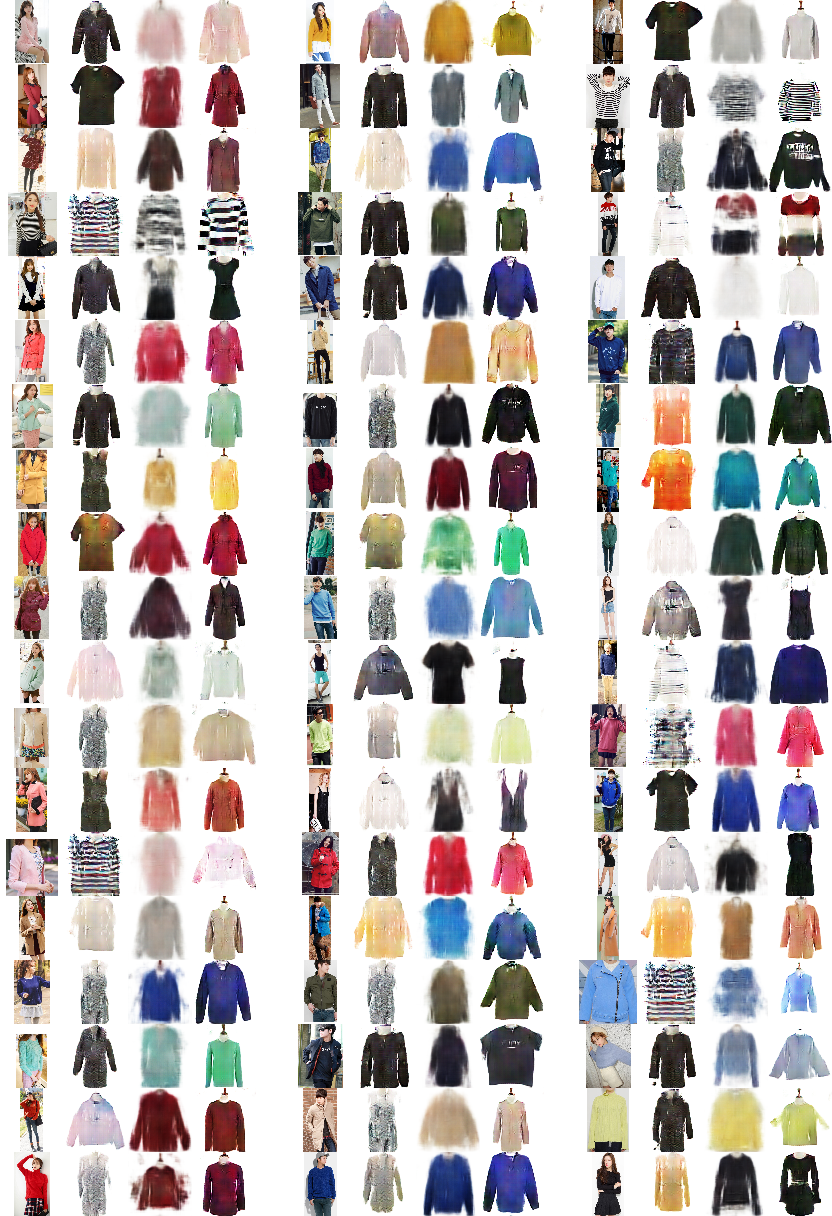}
\caption{Qualitative comparisons. Each image from the left to the right respectively corresponds to a source image, a ``C+RF'' result, a ``C+MSE'' result and our result. Each image is in 64$\times$64$\times$3 dimensions.}
\label{fig:comp}
\end{figure}

The left part of \Tref{tab:eval} shows the user study results. In the ``Realistic'' criteria, it is not surprising that ``C+MSE'' shows the worst performance due to the intrinsic limitation of the mean square loss for image generation. Its assumption of Gaussian distribution results in blurry images as shown in~\Fref{fig:comp}. However, the strong pairwise supervision of the mean square loss relatively succeeds in representing the category and attributes of a product.

When the converter is supervised with the real/fake discriminator only, the generated images are more realistic than those of ``C+MSE''. However, it fails to produce targets relevant to inputs and yields low attribute and category scores.

The user study results demonstrate the effectiveness of the proposed method. For all valuation criteria, our method outperforms the baselines. Especially, the ability to capture attributes and categories is better than that of ``C+MSE''. This result verifies the effectiveness of our domain discriminator. 

Another interesting observation is that our score of ``Realistic'' criteria is higher than that of ``C+RF''. Both of the methods include the real/fake discriminator but demonstrate distinct results. The difference may be caused by the domain discriminator which is added to the adversarial training in our method. When we train the domain discriminator, we regard all produced targets as ``unassociated''. This setting makes the the converter better transfer a source image into a more \textit{realistic} and relevant target image.

\subsection{Quantitative evaluation by pixel-level (dis)similarity}
\label{sec:quanti2}
For each method, we measure a pixel-level dissimilarity by Root Mean Square Error (RMSE) between a generated image and a target image over the test set. We also measure a pixel-level similarity by Structural Similarity (SSIM), since SSIM is known to be more consistent with human perception than RMSE. We use a color version of SSIM by averaging SSIMs for each channel.

The right part of \Tref{tab:eval} shows the results. As we can expect, ``C+MSE'' shows the lowest RMSE value because the converter is trained by minimizing the mean square loss. However, in case of SSIM, our method outperforms all the baselines.

To verify the effectiveness of the ``associated/unassociated'' supervision when we train the domain discriminator, we compare ours with ``C+RF+DD$-$Neg''. In \Tref{tab:eval}, our method outperforms this method. Without the irrelevant input pairs, the generation results could look realistic, but relatively fail to describe the attributes of items. This is why we added the irrelevant input pairs into supervision to encourage our model to capture discriminative attributes.

To verify the generalization capability of our model, we also compare ours with ``Retrieval by DD-score''. If our model fails in generalization (i.e. just memorizes and copies training items which are similar to query), our generation results could not be better than the retrieved items which are real. However, our method outperforms the retrieval method. It verifies the capability of our model to draw unseen items.


\begin{figure}[h!]
\footnotesize
\centering
\begin{tabular}{cc|cc|cc|cc|cc|cc}
Source&Ours&Source&Ours&Source&Ours&Source&Ours&Source&Ours&Source&Ours\\
\includegraphics[width=0.066\linewidth]{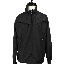}&\includegraphics[width=0.066\linewidth]{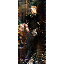}&\includegraphics[width=0.066\linewidth]{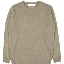}&\includegraphics[width=0.066\linewidth]{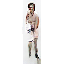}&\includegraphics[width=0.066\linewidth]{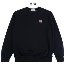}&\includegraphics[width=0.066\linewidth]{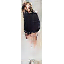}&\includegraphics[width=0.066\linewidth]{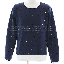}&\includegraphics[width=0.066\linewidth]{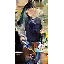}&\includegraphics[width=0.066\linewidth]{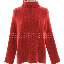}&\includegraphics[width=0.066\linewidth]{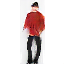}&\includegraphics[width=0.066\linewidth]{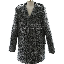}&\includegraphics[width=0.066\linewidth]{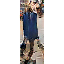}\\
\includegraphics[width=0.066\linewidth]{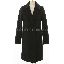}&\includegraphics[width=0.066\linewidth]{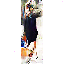}&\includegraphics[width=0.066\linewidth]{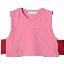}&\includegraphics[width=0.066\linewidth]{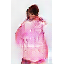}&\includegraphics[width=0.066\linewidth]{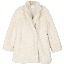}&\includegraphics[width=0.066\linewidth]{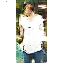}&\includegraphics[width=0.066\linewidth]{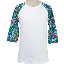}&\includegraphics[width=0.066\linewidth]{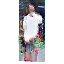}&\includegraphics[width=0.066\linewidth]{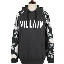}&\includegraphics[width=0.066\linewidth]{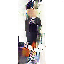}&\includegraphics[width=0.066\linewidth]{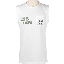}&\includegraphics[width=0.066\linewidth]{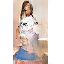}\\
\includegraphics[width=0.066\linewidth]{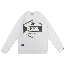}&\includegraphics[width=0.066\linewidth]{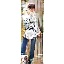}&\includegraphics[width=0.066\linewidth]{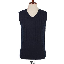}&\includegraphics[width=0.066\linewidth]{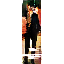}&\includegraphics[width=0.066\linewidth]{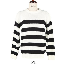}&\includegraphics[width=0.066\linewidth]{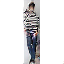}&\includegraphics[width=0.066\linewidth]{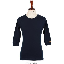}&\includegraphics[width=0.066\linewidth]{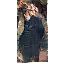}&\includegraphics[width=0.066\linewidth]{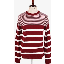}&\includegraphics[width=0.066\linewidth]{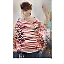}&\includegraphics[width=0.066\linewidth]{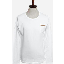}&\includegraphics[width=0.066\linewidth]{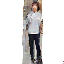}\\
\includegraphics[width=0.066\linewidth]{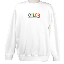}&\includegraphics[width=0.066\linewidth]{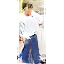}&\includegraphics[width=0.066\linewidth]{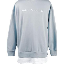}&\includegraphics[width=0.066\linewidth]{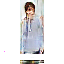}&\includegraphics[width=0.066\linewidth]{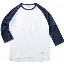}&\includegraphics[width=0.066\linewidth]{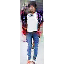}&\includegraphics[width=0.066\linewidth]{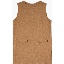}&\includegraphics[width=0.066\linewidth]{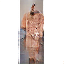}&\includegraphics[width=0.066\linewidth]{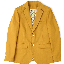}&\includegraphics[width=0.066\linewidth]{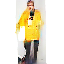}&\includegraphics[width=0.066\linewidth]{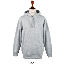}&\includegraphics[width=0.066\linewidth]{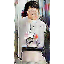}\\
\includegraphics[width=0.066\linewidth]{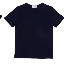}&\includegraphics[width=0.066\linewidth]{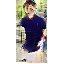}&\includegraphics[width=0.066\linewidth]{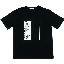}&\includegraphics[width=0.066\linewidth]{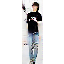}&\includegraphics[width=0.066\linewidth]{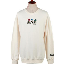}&\includegraphics[width=0.066\linewidth]{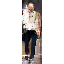}&\includegraphics[width=0.066\linewidth]{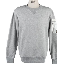}&\includegraphics[width=0.066\linewidth]{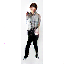}&\includegraphics[width=0.066\linewidth]{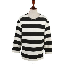}&\includegraphics[width=0.066\linewidth]{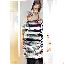}&\includegraphics[width=0.066\linewidth]{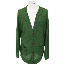}&\includegraphics[width=0.066\linewidth]{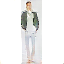}\\
\includegraphics[width=0.066\linewidth]{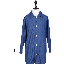}&\includegraphics[width=0.066\linewidth]{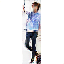}&\includegraphics[width=0.066\linewidth]{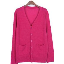}&\includegraphics[width=0.066\linewidth]{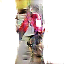}&\includegraphics[width=0.066\linewidth]{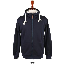}&\includegraphics[width=0.066\linewidth]{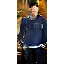}&\includegraphics[width=0.066\linewidth]{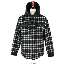}&\includegraphics[width=0.066\linewidth]{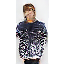}&\includegraphics[width=0.066\linewidth]{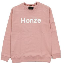}&\includegraphics[width=0.066\linewidth]{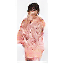}&\includegraphics[width=0.066\linewidth]{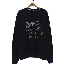}&\includegraphics[width=0.066\linewidth]{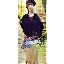}\\
\includegraphics[width=0.066\linewidth]{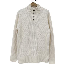}&\includegraphics[width=0.066\linewidth]{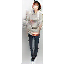}&\includegraphics[width=0.066\linewidth]{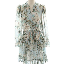}&\includegraphics[width=0.066\linewidth]{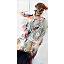}&\includegraphics[width=0.066\linewidth]{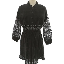}&\includegraphics[width=0.066\linewidth]{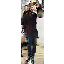}&\includegraphics[width=0.066\linewidth]{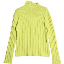}&\includegraphics[width=0.066\linewidth]{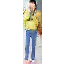}&\includegraphics[width=0.066\linewidth]{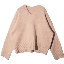}&\includegraphics[width=0.066\linewidth]{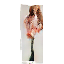}&\includegraphics[width=0.066\linewidth]{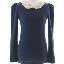}&\includegraphics[width=0.066\linewidth]{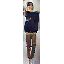}\\
\includegraphics[width=0.066\linewidth]{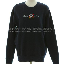}&\includegraphics[width=0.066\linewidth]{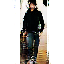}&\includegraphics[width=0.066\linewidth]{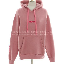}&\includegraphics[width=0.066\linewidth]{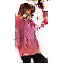}&\includegraphics[width=0.066\linewidth]{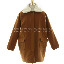}&\includegraphics[width=0.066\linewidth]{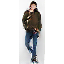}&\includegraphics[width=0.066\linewidth]{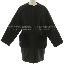}&\includegraphics[width=0.066\linewidth]{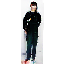}&\includegraphics[width=0.066\linewidth]{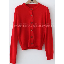}&\includegraphics[width=0.066\linewidth]{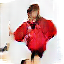}&\includegraphics[width=0.066\linewidth]{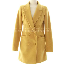}&\includegraphics[width=0.066\linewidth]{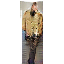}\\
\includegraphics[width=0.066\linewidth]{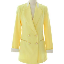}&\includegraphics[width=0.066\linewidth]{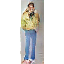}&\includegraphics[width=0.066\linewidth]{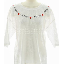}&\includegraphics[width=0.066\linewidth]{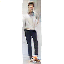}&\includegraphics[width=0.066\linewidth]{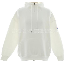}&\includegraphics[width=0.066\linewidth]{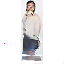}&\includegraphics[width=0.066\linewidth]{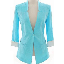}&\includegraphics[width=0.066\linewidth]{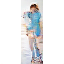}&\includegraphics[width=0.066\linewidth]{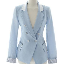}&\includegraphics[width=0.066\linewidth]{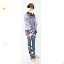}&\includegraphics[width=0.066\linewidth]{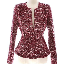}&\includegraphics[width=0.066\linewidth]{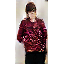}\\
\includegraphics[width=0.066\linewidth]{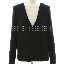}&\includegraphics[width=0.066\linewidth]{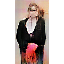}&\includegraphics[width=0.066\linewidth]{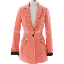}&\includegraphics[width=0.066\linewidth]{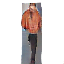}&\includegraphics[width=0.066\linewidth]{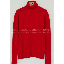}&\includegraphics[width=0.066\linewidth]{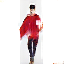}&\includegraphics[width=0.066\linewidth]{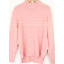}&\includegraphics[width=0.066\linewidth]{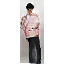}&\includegraphics[width=0.066\linewidth]{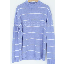}&\includegraphics[width=0.066\linewidth]{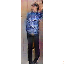}&\includegraphics[width=0.066\linewidth]{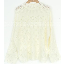}&\includegraphics[width=0.066\linewidth]{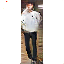}\\
\includegraphics[width=0.066\linewidth]{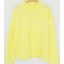}&\includegraphics[width=0.066\linewidth]{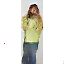}&\includegraphics[width=0.066\linewidth]{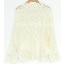}&\includegraphics[width=0.066\linewidth]{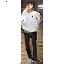}&\includegraphics[width=0.066\linewidth]{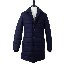}&\includegraphics[width=0.066\linewidth]{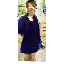}&\includegraphics[width=0.066\linewidth]{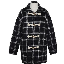}&\includegraphics[width=0.066\linewidth]{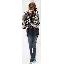}&\includegraphics[width=0.066\linewidth]{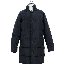}&\includegraphics[width=0.066\linewidth]{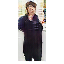}&\includegraphics[width=0.066\linewidth]{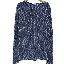}&\includegraphics[width=0.066\linewidth]{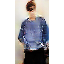}\\
\includegraphics[width=0.066\linewidth]{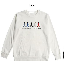}&\includegraphics[width=0.066\linewidth]{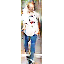}&\includegraphics[width=0.066\linewidth]{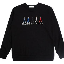}&\includegraphics[width=0.066\linewidth]{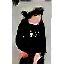}&\includegraphics[width=0.066\linewidth]{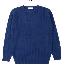}&\includegraphics[width=0.066\linewidth]{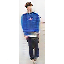}&\includegraphics[width=0.066\linewidth]{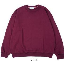}&\includegraphics[width=0.066\linewidth]{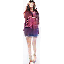}&\includegraphics[width=0.066\linewidth]{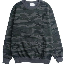}&\includegraphics[width=0.066\linewidth]{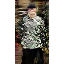}&\includegraphics[width=0.066\linewidth]{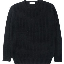}&\includegraphics[width=0.066\linewidth]{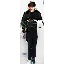}\\
\includegraphics[width=0.066\linewidth]{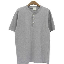}&\includegraphics[width=0.066\linewidth]{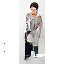}&\includegraphics[width=0.066\linewidth]{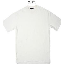}&\includegraphics[width=0.066\linewidth]{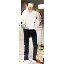}&\includegraphics[width=0.066\linewidth]{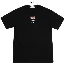}&\includegraphics[width=0.066\linewidth]{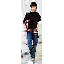}&\includegraphics[width=0.066\linewidth]{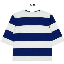}&\includegraphics[width=0.066\linewidth]{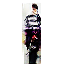}&\includegraphics[width=0.066\linewidth]{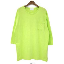}&\includegraphics[width=0.066\linewidth]{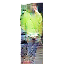}&\includegraphics[width=0.066\linewidth]{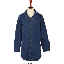}&\includegraphics[width=0.066\linewidth]{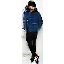}\\
\includegraphics[width=0.066\linewidth]{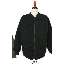}&\includegraphics[width=0.066\linewidth]{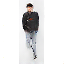}&\includegraphics[width=0.066\linewidth]{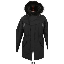}&\includegraphics[width=0.066\linewidth]{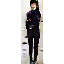}&\includegraphics[width=0.066\linewidth]{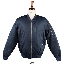}&\includegraphics[width=0.066\linewidth]{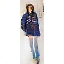}&\includegraphics[width=0.066\linewidth]{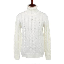}&\includegraphics[width=0.066\linewidth]{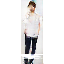}&\includegraphics[width=0.066\linewidth]{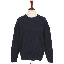}&\includegraphics[width=0.066\linewidth]{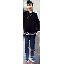}&\includegraphics[width=0.066\linewidth]{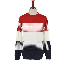}&\includegraphics[width=0.066\linewidth]{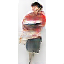}\\
\includegraphics[width=0.066\linewidth]{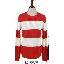}&\includegraphics[width=0.066\linewidth]{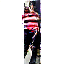}&\includegraphics[width=0.066\linewidth]{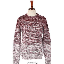}&\includegraphics[width=0.066\linewidth]{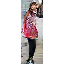}&\includegraphics[width=0.066\linewidth]{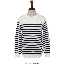}&\includegraphics[width=0.066\linewidth]{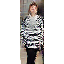}&\includegraphics[width=0.066\linewidth]{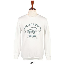}&\includegraphics[width=0.066\linewidth]{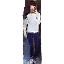}&\includegraphics[width=0.066\linewidth]{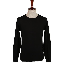}&\includegraphics[width=0.066\linewidth]{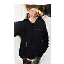}&\includegraphics[width=0.066\linewidth]{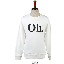}&\includegraphics[width=0.066\linewidth]{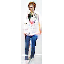}\\
\includegraphics[width=0.066\linewidth]{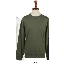}&\includegraphics[width=0.066\linewidth]{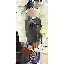}&\includegraphics[width=0.066\linewidth]{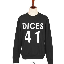}&\includegraphics[width=0.066\linewidth]{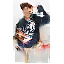}&\includegraphics[width=0.066\linewidth]{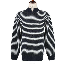}&\includegraphics[width=0.066\linewidth]{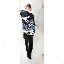}&\includegraphics[width=0.066\linewidth]{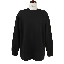}&\includegraphics[width=0.066\linewidth]{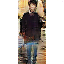}&\includegraphics[width=0.066\linewidth]{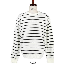}&\includegraphics[width=0.066\linewidth]{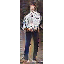}&\includegraphics[width=0.066\linewidth]{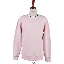}&\includegraphics[width=0.066\linewidth]{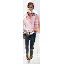}\\
\includegraphics[width=0.066\linewidth]{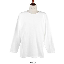}&\includegraphics[width=0.066\linewidth]{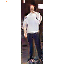}&\includegraphics[width=0.066\linewidth]{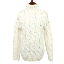}&\includegraphics[width=0.066\linewidth]{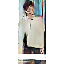}&\includegraphics[width=0.066\linewidth]{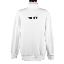}&\includegraphics[width=0.066\linewidth]{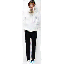}&\includegraphics[width=0.066\linewidth]{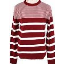}&\includegraphics[width=0.066\linewidth]{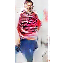}&\includegraphics[width=0.066\linewidth]{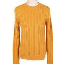}&\includegraphics[width=0.066\linewidth]{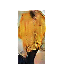}&\includegraphics[width=0.066\linewidth]{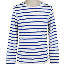}&\includegraphics[width=0.066\linewidth]{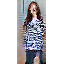}\\
\includegraphics[width=0.066\linewidth]{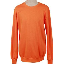}&\includegraphics[width=0.066\linewidth]{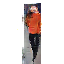}&\includegraphics[width=0.066\linewidth]{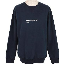}&\includegraphics[width=0.066\linewidth]{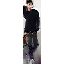}&\includegraphics[width=0.066\linewidth]{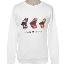}&\includegraphics[width=0.066\linewidth]{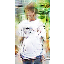}&\includegraphics[width=0.066\linewidth]{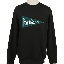}&\includegraphics[width=0.066\linewidth]{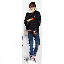}&\includegraphics[width=0.066\linewidth]{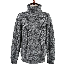}&\includegraphics[width=0.066\linewidth]{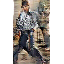}&\includegraphics[width=0.066\linewidth]{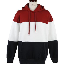}&\includegraphics[width=0.066\linewidth]{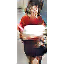}\\
\includegraphics[width=0.066\linewidth]{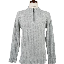}&\includegraphics[width=0.066\linewidth]{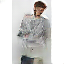}&\includegraphics[width=0.066\linewidth]{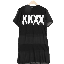}&\includegraphics[width=0.066\linewidth]{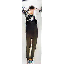}&\includegraphics[width=0.066\linewidth]{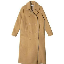}&\includegraphics[width=0.066\linewidth]{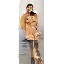}&\includegraphics[width=0.066\linewidth]{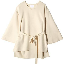}&\includegraphics[width=0.066\linewidth]{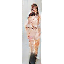}&\includegraphics[width=0.066\linewidth]{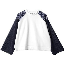}&\includegraphics[width=0.066\linewidth]{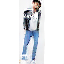}&\includegraphics[width=0.066\linewidth]{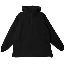}&\includegraphics[width=0.066\linewidth]{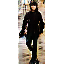}\\
\includegraphics[width=0.066\linewidth]{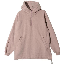}&\includegraphics[width=0.066\linewidth]{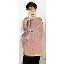}&\includegraphics[width=0.066\linewidth]{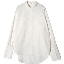}&\includegraphics[width=0.066\linewidth]{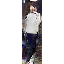}&\includegraphics[width=0.066\linewidth]{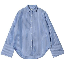}&\includegraphics[width=0.066\linewidth]{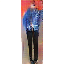}&\includegraphics[width=0.066\linewidth]{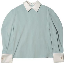}&\includegraphics[width=0.066\linewidth]{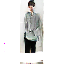}&\includegraphics[width=0.066\linewidth]{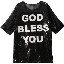}&\includegraphics[width=0.066\linewidth]{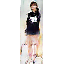}&\includegraphics[width=0.066\linewidth]{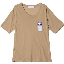}&\includegraphics[width=0.066\linewidth]{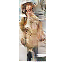}\\
\end{tabular}\\
\caption{100 chosen results of ``product to human''. Each image is shown in 64$\times$64$\times$3 dimensions.}
\label{fig:inverse}
\end{figure}

\Fref{fig:inverse} shows the results of \textit{``product to human''} setting. Since generating human is a more complex task, 65 epochs for initial training and 5 more epochs for fine-tuning are required for these results. All the other details are identical to those of the original setting. 
%\clearpage

\section{Conclusion}
We have presented pixel-level domain transfer based on Generative Adversarial Nets framework. The proposed domain discriminator enables us to train the semantic relation between the domains, and the converter has succeeded in generating decent target images. Also, we have presented a large dataset that could contribute to domain adaptation researches. Since our framework is not constrained to specific problems, we expect to extend it to other types of pixel-level domain transfer problems from low-level image processing to high-level synthesis.

\clearpage

\bibliographystyle{splncs03}
\bibliography{egbib}
\end{document}